\documentclass[letterpaper]{article} 
\usepackage{aaai25}  
\usepackage{times}  
\usepackage{helvet}  
\usepackage{courier}  
\usepackage[hyphens]{url}  
\usepackage{graphicx} 
\usepackage{natbib}  
\usepackage{caption} 
\usepackage{booktabs} 
\usepackage{multirow} 
\usepackage{makecell} 
\usepackage{siunitx} 
\usepackage{graphicx} 
\usepackage{subcaption} 
\usepackage{array} 
\usepackage{booktabs} 
\usepackage{amsmath} 
\usepackage{amsfonts} 
\usepackage{amssymb} 
\usepackage{algorithm}
\usepackage{algorithmic}

\usepackage{newfloat}
\usepackage{listings}
\DeclareCaptionStyle{ruled}{labelfont=normalfont,labelsep=colon,strut=off} 
\floatstyle{ruled}
\newfloat{listing}{tb}{lst}{}
\floatname{listing}{Listing}
\title{GeoAggregator: An Efficient Transformer Model for Geo-Spatial Tabular Data}
\author{
    Rui Deng\textsuperscript{\rm 1}, Ziqi Li\textsuperscript{\rm 2}, Mingshu Wang\textsuperscript{\rm 1}\thanks{Corresponding author}\\
}
\affiliations{
    \textsuperscript{\rm 1}School of Geographical and Earth Science, University of Glasgow\\

    \textsuperscript{\rm 2}Department of Geography, Florida State University\\
    \{rui.deng, mingshu.wang\}@glasgow.ac.uk, ziqi.li@fsu.edu
}

\begin{document}

\maketitle

\begin{abstract}
Modeling geospatial tabular data with deep learning has become a promising alternative to traditional statistical and machine learning approaches. However, existing deep learning models often face challenges related to scalability and flexibility as datasets grow. To this end, this paper introduces GeoAggregator, an efficient and lightweight algorithm based on transformer architecture designed specifically for geospatial tabular data modeling. GeoAggregators explicitly account for spatial autocorrelation and spatial heterogeneity through Gaussian-biased local attention and global positional awareness. Additionally, we introduce a new attention mechanism that uses the Cartesian product to manage the size of the model while maintaining strong expressive power. We benchmark GeoAggregator against spatial statistical models, XGBoost, and several state-of-the-art geospatial deep learning methods using both synthetic and empirical geospatial datasets. The results demonstrate that GeoAggregators achieve the best or second-best performance compared to their competitors on nearly all datasets. GeoAggregator's efficiency is underscored by its reduced model size, making it both scalable and lightweight. Moreover, ablation experiments offer insights into the effectiveness of the Gaussian bias and Cartesian attention mechanism, providing recommendations for further optimizing the GeoAggregator's performance.
\end{abstract}

%
\begin{links}
    \link{Code and Data}{https://github.com/ruid7181/GeoAggregator}
\end{links}

\section{Introduction}

Geospatial data are increasingly available with the widespread deployment of GPS-enabled sensors and the growing demand for location-based services \cite{luo2021stan, stewart2022torchgeo}. Geospatial data modeling is crucial in natural and social sciences to understand intricate spatial relationships, predict future spatial scenarios, and inform decision-making. Its applications span various fields, including public health, environmental science, urban and regional planning, among others \cite{karimi2017geospatial}. To date, geospatial studies have primarily relied on spatial and geo-statistical models. Although these models explicitly account for spatial effects, such as spatial autocorrelation (SA) and spatial heterogeneity (SH), they often depend on strong data and model assumptions. Besides, their effectiveness diminishes when handling large volumes of data or modeling complex non-linearity due to their high computational complexity and linear nature \cite{li2022extracting}.

In addition to classic methods, efforts have been broadened to cover a wide range of deep learning paradigms \cite{jiang-2019-spa-predict-survey}. By explicitly defining a graph to represent the data points and the underlying spatial structure, the problems can be approximated to node-level learning tasks. Graph-based networks can be trained for node regression and classification \cite{zhu2020understanding, Wu-graph-conv-regression, zhu2022spatial}. Alternatively, with a spatial proxy grid proposed by \cite{DaiPm25CNN}, convolutional networks can work on irregular grids and effectively model SH in spatial regression. On top of these approaches, ensemble learning methods have also been applied to further improve predictive performance \cite{cheng2024ensemble}. However, the above models are typically associated with large input sizes and a considerable number of parameters, which makes them scale poorly on large datasets.

Lately, transformer models have demonstrated their effectiveness in handling irregular grids \cite{lee2024inducing}, point clouds \cite{zhao2021point} as well as geospatial datasets \cite{li2023ssin, he2023hierarchical, jia2024geotransformer, unlu2024geotokens}. The attention mechanism selectively focuses on elements in a sequence and aggregates the information according to relation-based attention scores. This approach brings new horizons for geospatial data modeling in that it accounts for contextual information based on proximity, and is sensitive to positional information \cite{vaswani2017attention, su2024roformer, vivanco2024geoclip}. That is, it allows information aggregation from neighboring points to learn SA patterns while remaining aware of SH patterns across the global space. Besides, transformers represent a group of modern architectures that can serve as a general-purpose feature extractor for multiple tasks and multi-modal fusion \cite{jaegle2021perceiver, xu2023multimodal}.

However, the well-known problem of computational and time complexity increasing quadratically with input sequence length limits the application of transformers, especially in geospatial contexts, where the size of real-world datasets varies dramatically in different scenarios. To overcome this issue, efforts have focused on optimizing the self-attention mechanism and transformer architecture (e.g., \cite{jaegle2021perceiver, dao2022flashattention}). Nevertheless, efficiency improvements in the context of geospatial tabular data remain limited. Moreover, most models do not explicitly incorporate geographical priors to better account for geospatial effect,s including SA and SH. In this work, we propose GeoAggregator, an efficient and lightweight transformer model enhanced by geographical priors for geospatial tabular data modeling. We demonstrate the effectiveness of GeoAggregator in various geospatial regression tasks.

\section{Background}
\subsection{Modeling Spatial Effects}
SA and SH are two main spatial effects that govern the distribution and interaction of spatial data \cite{30yearsAnselin}. SA refers to the phenomenon where spatial data exhibit spatial clustering, meaning that similar values are observed in close geographic proximity. On the other hand, SH refers to the differences in spatial patterns and processes at different locations. This means that relationships or patterns observed in one place may not be the same in another. These differences are often due to local contexts, such as varying environmental and socio-economic conditions, and cultural and policy influences that are hard to measure or quantify \cite{fotheringham2023measuring}.

One common approach to express SA is to use spatially lagged variables to capture the dependency with nearby values when modeling a target location. One can formally assign a spatial weight between pairs of {$N$} locations to build a spatial weight matrix {$\mathbf{W} \in \mathbb{R}^{N \times N}$} to characterize the degree of spatial interactions. Then, a spatial lag term can be expressed as:
\begin{align}
    \tilde{\mathbf{y}} = \rho \mathbf{W} \mathbf{y}
\end{align}
where {$\mathbf{y} \in \mathbb{R}^N$} is a vector containing {$N$} data points, {$\rho$} is a parameter to control the strength of SA. It can be seen that determining optimized {$\mathbf{W}$} and {$\rho$} is crucial for properly expressing SA.

SH is often addressed using \textit{local} modeling approaches. The most well-known example is Geographically Weighted Regression (GWR), which allows regression parameters to be location-specific \cite{fotheringham2003geographically}:
\begin{align}
    y_i = \mathbf{x}_i \boldsymbol{\beta}_i + \epsilon_{i}
\end{align}
where {$i=0,1,\ldots,N-1$}; {$\boldsymbol{\beta}_i \in \mathbb{R}^{p}$} is the corresponding {$p$} regression coefficients, which can be estimated by:
\begin{align}
    \hat{\boldsymbol{\beta}}_i = [\mathbf{X}^T\mathbf{W}_i\mathbf{X}]^{-1} \mathbf{X}^T\mathbf{W}_i\mathbf{Y}
\end{align}
where {$\mathbf{X} = [\mathbf{x}_1^T,\mathbf{x}_2^T,\ldots,\mathbf{x}_N^T]^T$}, {$\mathbf{W}_i = diag(w_{i,1}, w_{i,2}, \ldots, w_{i,N})$} is a spatial weight matrix. Note that {$w_{i,j}$} can be given by a stationary kernel function such as:
\begin{align}
    w_{i,j} = K(d_{i,j}) = e^{-\frac{1}{2}(\frac{d_{i,j}}{\gamma})^2}
\end{align}
based on the distance {$d_{i,j}$} between two points, scaled by a bandwidth {$\gamma$}.

\begin{figure*}[h]
    \centering
    \begin{subfigure}[b]{\textwidth}
        \centering
        \includegraphics[width=0.95\textwidth]{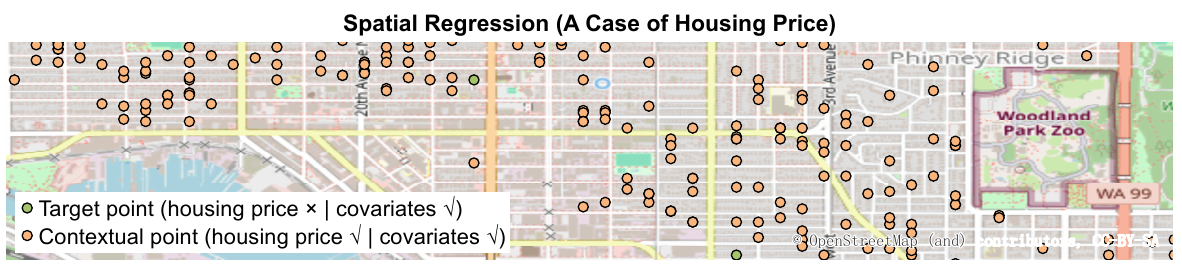}
    \end{subfigure}
    \begin{subfigure}[b]{\textwidth}
        \centering
        \includegraphics[width=\textwidth]{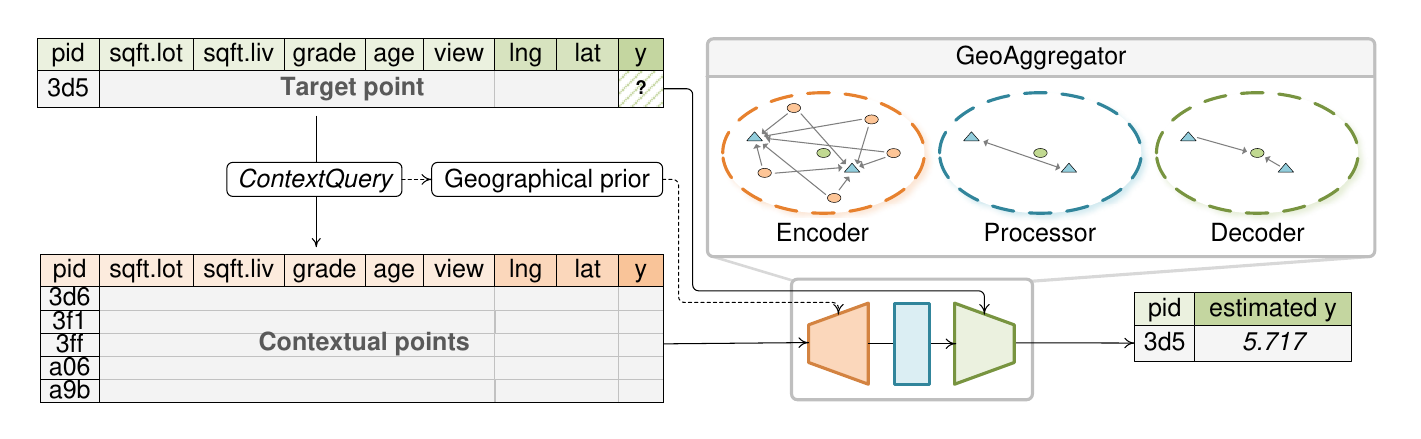}
    \end{subfigure}
    \caption{An illustration of the geospatial regression workflow. Each data point has several covariates and a spatial location. The target variable (the housing price in this case) is observed only for \textit{part} of the points. Using aggregated neighborhood information, we propose an encoder-processor-decoder architecture to predict unobserved target variables.}
    \label{fig:combined}
\end{figure*}

\subsection{Problem Statement}
We consider the following geospatial regression problem. Given a set {$\mathcal{P}_c$} of data points in a 2D continuous geographical space. Each point (referred to as contextual point) is located by a 2D spatial location vector {$\mathbf{l}^c \in \mathbb{R}^2$} while associated with {$m$} covariates organized as {$\mathbf{x}^c \in \mathbb{R}^m$} and a target variable {$y^c \in \mathbb{R}$}. We denote the {$i$}-th contextual points as {$\boldsymbol{p}^c_i = (\mathbf{x}^c_i, \mathbf{l}^c_i, y^c_i)$}. For another set {$\mathcal{P}_t$} of data points in the same 2D space, {$m$} covariates {$\mathbf{x}^t \in \mathbb{R}^m$} are observed, but the target variable {$y^t$} is missing, i.e., {$\mathcal{P}_t=\{\boldsymbol{p}^t_j\}=\{(\mathbf{x}^t_j, \mathbf{l}^t_j)\}$}. Note that a point set can be organized in a tabular format with {$\left| \mathcal{P} \right|$} rows and {$m + 3$} attributes.

We define a mapping {$\textit{ContextQuery}: \mathcal{P}_t \to \mathcal{N}_r(\mathcal{P}_c)$}, where \( \mathcal{N}_r \subseteq \mathcal{S}(\mathcal{P}_c) \) denotes a collection of subsets of {$\mathcal{P}_c$} determined by spatial proximity. For each {$\boldsymbol{p}^t \in \mathcal{P}_t$}, the mapping is a subset of {$\mathcal{P}_c$} consisting of neighboring points of {$\boldsymbol{p}^t$} within a query radius {$r$}.

We assume target variable {$y$} is generated by unknown processes that potentially exhibit a mixture of SA and SH effects. The aim is to learn to predict the unobserved target variable {$y^t_j$} of {$\boldsymbol{p}^t_j$} as a function of 1) covariates and spatial location of {$\boldsymbol{p}^t_j$}; and 2) covariates, spatial location and target variable of all {$\boldsymbol{p}^c_i \in \textit{ContextQuery}(\boldsymbol{p}^t_j)$}.

\section{Approaches}
This section introduces the proposed GeoAggregator model for geospatial regression tasks. Figure 2 provides a high-level overview of the architecture.

\subsection{Input Sequence}
GeoAggregator takes in a sequence of points with a maximum length of {$\ell_{max}$}. For each target point {$\boldsymbol{p}^t_j \in \mathcal{P}_t$}, we use the set {$\boldsymbol{h}^t_j=\textit{ContextQuery}(\boldsymbol{p}^t)$} as the corresponding input sequence and adjust the length of the sequence by clipping and padding to match {$\ell_{max}$}:
\begin{align}
    \boldsymbol{h}^{in}_j =
    \begin{cases}
        (\boldsymbol{h}^t_j)_{0:\ell_{max}} & \text{if } \left| \boldsymbol{h}^t_j \right| > \ell_{max} \\
        \textrm{ZeroPad}_{\ell_{max}}(\boldsymbol{h}^t_j) & \text{if } \left| \boldsymbol{h}^t_j \right| \leq \ell_{max} 
    \end{cases}
\end{align}
where {$\textrm{ZeroPad}_{\ell_{max}}(\cdot)$} is the operation to pad a sequence with 0 to the length of {$\ell_{max}$}. Note that the clipping operation randomly removes contextual points since the order of the points in {$\boldsymbol{h}^{t}_j$} is random.

We then generate a mask sequence {$\boldsymbol{m}^{in}$} for each target point {$\boldsymbol{p}^t_j$} to ensure that the encoder attends only to the non-zero points:
\begin{align}
    m_k^{in} = 
    \begin{cases} 
        0 & \text{if } 0 \leq k < \min(\left| \boldsymbol{h}^t_j \right|, \ell_{max} - 1) \\
        1 & \text{if } \min(\left| \boldsymbol{h}^t_j \right|, \ell_{max} - 1) \leq k < \ell_{max}
    \end{cases}
\end{align}
We denote the actual number of contextual points in the input sequence as {$\ell_{in}=\ell_{max} - \sum_{\underset{}{k}} m_k^{in}$}.

\subsection{Feature Projection}

We first use a slightly modified batch normalization to normalize each covariate within a mini-batch \cite{bjorck2018understanding}. Note that we only use unmasked points to calculate the mean/standard deviation values.

We use two parallel dense networks with {$\textrm{Tanhshrink}$} activation for feature projection. Covariates ({$\mathbf{x}^c$} and {$\mathbf{x}^t$}) and the target variable ({$y^c$}) are projected into higher dimensional feature space separately. This can be written as:
\begin{align}
    \mathbf{e}_x=\textrm{Dense}_x(\mathbf{x}^{\{t,c\}}) \in \mathbb{R}^{{\frac{d_{model}}{2}}}
\end{align}
\begin{align}
    \mathbf{e}_y=\textrm{Dense}_y(y^c) \in \mathbb{R}^{{\frac{d_{model}}{2}}}
\end{align}
where {$d_{model}$} is the dimension of the latent embedding in the GeoAggregator model. Since {$y^t$} is unknown, we use a learnable feature vector instead.

\subsection{2D Rotary Positional Embedding}
Due to the inherent permutation invariance, positional information must be explicitly injected into the attention mechanism. Inspired by \cite{su2024roformer}, \cite{mai2020multi}, and \cite{unlu2024geotokens}, we use an augmented rotation matrix to incorporate 2D spatial locations into the embedding features.

We first propose the sinusoidal representation features for this purpose. For a data point {$\boldsymbol{p}$} at the spatial location {$\mathbf{l}=(l_{1}, l_{2})$}, the representation feature at scale {$s$} is {$\mathcal{\varphi}_{s}=\{\cos{(l_{1}\theta_s)},\sin{(l_{1}\theta_s)},cos{(l_{2}\theta_s)},\sin{(l_{2}\theta_s)}\}$}, where {$\theta_s=10000^{\frac{2-2s}{d}}$}; {$s=0,1,\ldots,S-1$}; {$S=\frac{d}{4}$} and {$d$} is some embedding dimension. We construct {$4$}-by-{$4$} 2D rotation matrices based on {$\mathbf{\Phi}_s$} that rotate the embedding feature vectors alternatively in the two dimensions of {$\mathbf{l}$}:
\begin{align}
    \scriptsize
\mathbf{\Phi}_{s}=
\begin{pmatrix}
\cos{(l_{i,1}\theta_s)} & -\sin{(l_{i,1}\theta_s)} & 0 & 0 \\
\sin{(l_{i,1}\theta_s)} & \cos{(l_{i,1}\theta_s)} & 0 & 0 \\
0 & 0 & \cos{(l_{i,2}\theta_s)} & -\sin{(l_{i,2}\theta_s)} \\
0 & 0 & \sin{(l_{i,2}\theta_s)} & \cos{(l_{i,2}\theta_s)} \\
\end{pmatrix}
\end{align}
The {$d$}-by-{$d$} rotation matrix can then be constructed by placing the rotation matrices at each scale along the diagonal:
\begin{align}
\mathbf{\Phi}=diag(\mathbf{\Phi}_{0},\mathbf{\Phi}_{1},\cdots,\mathbf{\Phi}_{S}) \in \mathbb{R}^{d \times d}
\end{align}
We inject the spatial location information into embedding features produced in section 3.3 through a matrix multiplication \cite{su2024roformer}:
\begin{align}
    \tilde{\mathbf{e}} = \mathbf{\Phi} \mathbf{e}
\end{align}
\subsection{Cartesian Product Attention}
The learnable parameters in attention are mainly contributed by the linear projections, which produce Query (Q), Key (K), and Value (V) vectors with rich information. Consequently, as the embedding dimension and the number of layers increase, the model size scales rapidly. This poses challenges for even moderately sized geospatial datasets. To address this, we propose Multi-head Cartesian Product Attention (MCPA)  to help manage the learnable parameters and computational complexity while maintaining the expressive power as much as possible.

As shown in Figure 2, our MCPA takes two input embeddings from the feature projection: {$\tilde{\mathbf{e}}_x$} from covariates and {$\tilde{\mathbf{e}}_y$} from the target variable. Like in the feature projection, we employ parallel linear projections {$\mathbf{W}^{\{Q_x,K_x,V_x\}}_{hx} \in \mathbb{R}^{\frac{d_{model}}{2} \times {d_c}}$} and {$\mathbf{W}^{\{Q_y,K_y,V_y\}}_{hy} \in \mathbb{R}^{\frac{d_{model}}{2} \times d_c}$} to map {$\tilde{\mathbf{e}}_x$} and {$\tilde{\mathbf{e}}_y$} to a {$\mathbb{R}^{d_c}$} space. Note that {$hx, hy = 0, 1, \ldots, H$} and the following equation holds: {$2d_c \cdot H^2 = d_{model}$}. 

Here, for each point, we define two sets of the mapped embeddings in the {$\mathbb{R}^{d_c}$} space: {$\mathcal{X}$} and {$\mathcal{Y}$}, with {$\left| \mathcal{X} \right| = \left| \mathcal{Y} \right| = H$}. Cartesian product is conducted between them as a concatenation strategy to generate new embedding sets:
\begin{align}
    \mathcal{X} \times \mathcal{Y} = \{\mathbf{e}_c=[\mathbf{x}; \mathbf{y}] \mid \mathbf{x} \in \mathcal{X} \text{ and } \mathbf{y} \in \mathcal{Y}\}
\end{align}
where {$[;]$} is the concatenation operation to enrich the feature interaction while ensuring the attention outputs are still in the same dimensions as the inputs. {$\mathbf{e}_c \in \mathbb{R}^{2 d_c}$} and {$\left| \mathcal{X} \times \mathcal{Y} \right| = H^2$} is the total number of heads.

After the attention operation, we rearrange the orders to reform the embeddings of covariates and the target variable (indicated in the right of Figure 2). Finally, two linear projections {$\mathbf{W}^{\{O_x,O_y\}} \in \mathbb{R}^{\frac{d_{model}}{2} \times \frac{d_{model}}{2}}$} are used to map the rearranged features into a final output.

\subsection{Gaussian Attention Bias}
Attention mechanisms intrinsically model internal interactions of a sequence, i.e., they selectively aggregate information that is beneficial to the task. However, the vanilla attention does not explicitly consider the effect of spatial proximity.

To tackle this problem, we propose to bias the interactions directly with a geographical prior. The Gaussian kernel, widely used in GWR models, is a common choice in various applications (as given in Equation 4) \cite{jia2024geotransformer}. We, therefore, adjust the attention coefficients with a Gaussian bias, derived from the spatial proximity of point pairs \cite{Guo_2019_Gaussian-tranasformer, kim2023understanding}. Given the embeddings of these two points, {$\mathbf{e}_i$} and {$\mathbf{e}_j$}:

\begin{align}
    \alpha_{i,j} = \frac{\textrm{exp}(\mathbf{e}_i \cdot \mathbf{e}_j)}{\sum_{\underset{}{l}} \textrm{exp}(\mathbf{e}_i \cdot \mathbf{e}_l)}
\end{align}

\begin{equation}
    \begin{split}
        \tilde{\mathbf{e}}_i &= \sum_{\underset{}{j}} \frac{\textrm{exp}(-d^2_{i,j}) \cdot \alpha_{i,j}}{\sum_{\underset{}{k}} \textrm{exp}(-d_{i,k}^2) \cdot \alpha_{i,k}} \cdot \mathbf{e}_j \\
        &= \sum_{\underset{}{j}} \frac{\textrm{exp}(\mathbf{e}_i \cdot \mathbf{e}_j - d^2_{i,j})}{\sum_{k} \textrm{exp}(\mathbf{e}_i \cdot \mathbf{e}_k - d^2_{i,k})} \cdot \mathbf{e}_j \\
        &= \sum_{\underset{}{j}} \textrm{softmax}(\mathbf{e}_i \cdot \mathbf{e}_j - d^2_{i,j}) \cdot \mathbf{e}_j
    \end{split}
\end{equation}
where {$\alpha_{i,j}$} denotes the attention coefficient between {$\mathbf{e}_i$} and {$\mathbf{e}_j$}. We add a learnable attention bias factor {$\lambda \in \mathbb{R}$} to control the magnitude of the bias so that:
\begin{align}
    \tilde{\mathbf{e}}_i &= \sum_{\underset{}{j}} \textrm{softmax}(\mathbf{e}_i \cdot \mathbf{e}_j - \lambda d^2_{i,j}) \cdot \mathbf{e}_j
\end{align}

The effectiveness of the proposed Gaussian bias is examined in the ablation study.

\begin{figure*}
    \centering
    \includegraphics[width=1\linewidth]{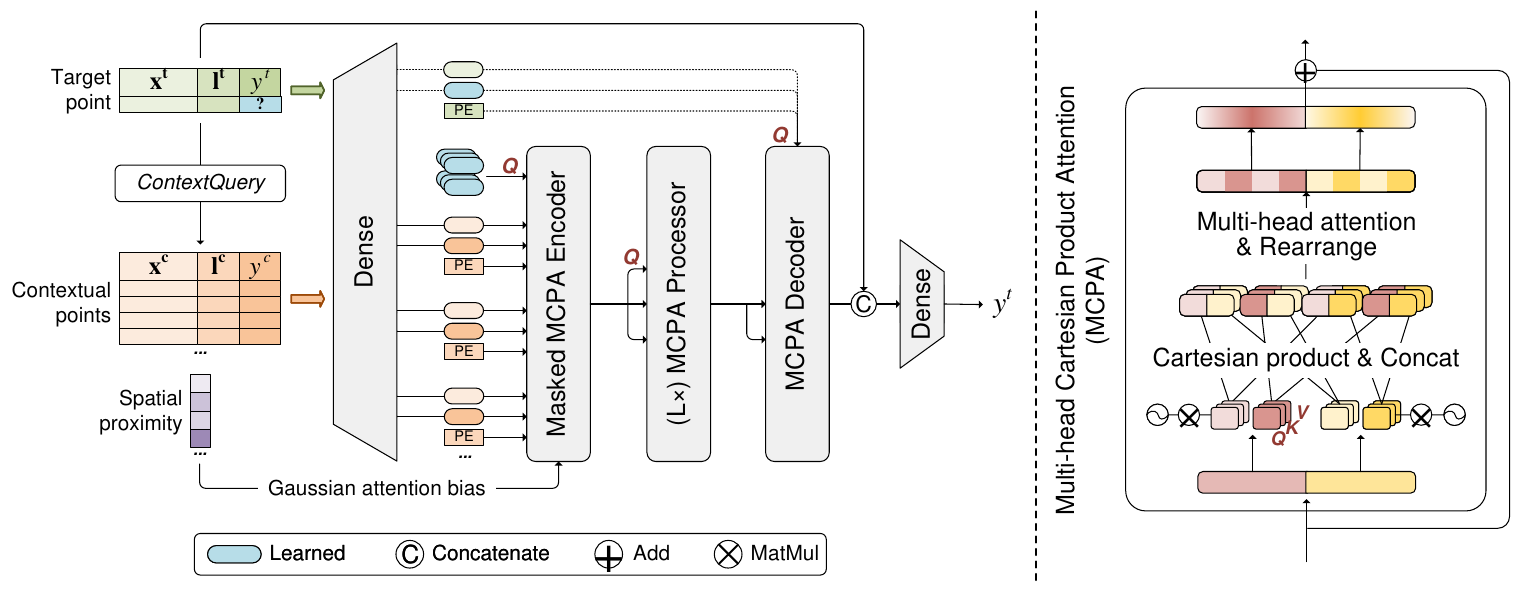}
    \caption{Illustration of the GeoAggregator model and the Multi-head Cartesian Product Attention (MCPA) mechanism.}
    \label{fig:enter-label}
\end{figure*}

\subsection{GeoAggregator Model}
Overall, GeoAggregator has an \textit{encoder-processor-decoder} architecture that achieves end-to-end prediction.
\subsubsection{Encoder.}
The encoder contains one MCPA operation, compressing the information from the contextual points into {$\ell_{hidden}$} learnable states (denoted as inducing points). We inject the spatial location of the target point into the inducing points. This enables each inducing point to learn different interaction patterns between contextual and target locations.
\subsubsection{Processor.}
Similar to \cite{lee2024inducing}, we introduce {$L$} processor modules so that the computational complexity of all-point attention ({$O(\ell_{in}^2)$}) is eased to {$O(\ell_{in} \cdot \ell_{hidden})$}, where {$\ell_{hidden} \ll \ell_{in}$}. Therefore, the computational complexity grows linearly with the input sequence length, as will be shown in later analysis. The processor module learns the interactions between {$\ell_{hidden}$} inducing points.
\subsubsection{Decoder and Output Head.}
The decoder is a cross-attention module that queries inducing points using the target point. We append a dense network with {$\textrm{Tanhshrink}$} activation as the output head. We concatenate the decoded embeddings and the original features {$\mathbf{x}_i^t$} and {$\mathbf{l}_i^t$} to ensure that a linear solution is always a subclass of our model.

\section{Experiments}
\subsubsection{Datasets.}
Experiments are first conducted on eight synthetic datasets to test the performance of models in handling different spatial processes. We generate the datasets using four different spatial processes: \textbf{Linear Model} (Lin) \cite{li2022extracting}, \textbf{Spatial Lagged Model} (SL), \textbf{Spatial Lagged X Model} (SLX) \cite{anselin2009spatial}, and \textbf{Spatial Durbin Model} (Durbin) \cite{mur2005closer}. Additionally, two types of covariates ({$\mathbf{x}$}) are generated: 1) sampled from a uniform distribution ({$x \sim U(-1,1)$}) whose values are independent of spatial location (denoted as \{Lin, SL, SLX, Durbin\}-r), and 2) sampled from a real-world Digital Elevation Map (DEM) which exhibits spatial autocorrelation (denoted as \{Lin, SL, SLX, Durbin\}-d).

To further illustrate the performances, we use three real-world datasets of different sizes: (1) \textbf{PM25}: from \cite{DaiPm25CNN}, which includes 1,457 PM2.5 concentration measurements across mainland China, coupled with related environmental factors. It presents a long-range spatial regression problem due to its sparse and uneven data distribution; 
(2) \textbf{Housing}: as per \cite{li2024geoshapley}, this dataset contains housing prices in King County, WA, USA. It represents a small-scale, densely distributed spatial dataset with notable spatial effects;
(3) \textbf{Poverty}: Sourced from \cite{MaryniaSDOH}, this dataset includes 14 socioeconomic variables that estimate poverty levels across the continental US. It features a mid-range spatial regression challenge with densely distributed neighborhood-level data.

\begin{table}
    \centering
    \small
    \setlength{\tabcolsep}{2pt}
    \begin{tabular}{cccc}
    \toprule
        Dataset & \#data point & \#covariate & \makecell{range of \\ spatial effect} \\
        \hline
        Synthetic datasets & 2500 & 2 & -\\
        PM25 & 1457 & 7 & mid-range\\
        Housing & 16580 & 8 & short-range\\
        Poverty & 71900 & 14 & long-range\\
    \bottomrule
    \end{tabular}
    \caption{Basic profiles of the datasets used in this paper.}
    \label{tab:my_label}
\end{table}
We set the splitting ratio of training-validation-testing to be 7:1:2 for all datasets except for the PM25 dataset, whose splitting ratio is 56:14:30 \cite{DaiPm25CNN}.

\subsubsection{Baseline Models.}
We select five models as baselines. One classic spatial statistical model, Geographically Weighted Regression (\textbf{GWR}) \cite{brunsdon1998geographically}. One tree-based ensemble model, \textbf{XGBoost} \cite{chen2016xgboost}. We also select three representative deep networks, Spatial Regressive Graph Convolutional Network (\textbf{SRGCNN}) \cite{zhu2022spatial}, Geographically CNN Weighted Regression (\textbf{GCNNWR}) \cite{DaiPm25CNN}, and Geographical Spatial Heterogeneous Ensemble Learning model (\textbf{GSH-EL}) \cite{cheng2024ensemble}. Additionally, we include 3 GeoAggregator models with the MCPA operation replaced by vanilla attention (denoted as \textbf{Vanilla} in the following experiments) to demonstrate the effectiveness of the MCPA mechanism.

\subsubsection{Evaluation Metrics.}
We use mean absolute error (MAE) as an objective function and evaluation metric and report the R-square ({$R^2$}) as another evaluation metric. We report the mean value from 3 repeated training and testing runs.

\subsubsection{Implementation and Training Details.}
We set the latent dimension {$d_{model}$} as 32, total number of heads {$H^2=4$}. We implement three versions of the GeoAggregator, with the number of processor modules {$L=0, 1, 2$} named GeoAggregator-mini (GA-mini), GA-small, and GA-large, respectively. We set {$l_{hidden}$} to be 0, 4, 8; the parameter {$l_{max}$} to be 81, 144, and 256, respectively. Corresponding searching radius in the {$ContextQuery$} operation is estimated on training datasets.

We use the Adam optimizer with a cyclical learning rate scheduler (max learning rate is {$5 \times 10^{-3}$}) \cite{kingma2014adam, popel2018training}. We conduct synthetic data experiments on a laptop with 32GB of RAM and real-world data experiments on a Google Colab virtual machine equipped with an NVIDIA P100 GPU with 16GB of GPU memory.

\subsection{Performance on Synthetic Datasets}
The regression performances on eight synthetic datasets are listed in Table 2. 3 GeoAggregator models exhibit the strongest or most competitive performance. GA-mini performs the best in all six variants of transformer models, achieving four best results. This showcases its strong ability to capture stationary/non-stationary spatial patterns and to learn global-scale SH. For the convolutional solutions, the GCNNWR model achieves four best results, the same as our GA-mini. The competitive performances indicate that GCNNWR learns to map the spatial proximity grid directly to regression coefficients. SRGCNN and GSH-EL perform less satisfactorily on datasets with randomly sampled covariates. This means they overemphasize the global point interactions, which introduce noise instead of useful information. Two non-deep learning models, GWR and XGBoost, exhibit strong robustness across all datasets, although they are not among the top solutions. 

Generally, more detailed hyperparameter tuning of the GeoAggregator, including varying {$l_{hidden}$}, {$L$}, and {$l_{max}$}, could potentially lead to better results. Besides, larger GeoAggregator models tend to perform worse than smaller ones. This could be due to increased overfitting and difficulty in model training.

\begin{table*}
    \centering
    \small
    \renewcommand{\arraystretch}{0.95}
    \setlength{\tabcolsep}{2pt}
    \begin{tabular}{crrrrrrrrrrrrrrrrc}
    \toprule
         \multirow{2}*{Model} & \multicolumn{2}{c}{Lin-r} & \multicolumn{2}{c}{SL-r} & \multicolumn{2}{c}{SLX-r} & \multicolumn{2}{c}{Durbin-r} & \multicolumn{2}{c}{Lin-d} & \multicolumn{2}{c}{SL-d} & \multicolumn{2}{c}{SLX-d} & \multicolumn{2}{c}{Durbin-d} & \multirow{2}*{\#best} \\
         \cmidrule(lr){2-3}
         \cmidrule(lr){4-5}
         \cmidrule(lr){6-7}
         \cmidrule(lr){8-9}
         \cmidrule(lr){10-11}
         \cmidrule(lr){12-13}
         \cmidrule(lr){14-15}
         \cmidrule(lr){16-17}
         ~ & \multicolumn{1}{c}{MAE} & \multicolumn{1}{c}{$R^2$} & \multicolumn{1}{c}{MAE} & \multicolumn{1}{c}{$R^2$} & \multicolumn{1}{c}{MAE} & \multicolumn{1}{c}{$R^2$} & \multicolumn{1}{c}{MAE} & \multicolumn{1}{c}{$R^2$} & \multicolumn{1}{c}{MAE} & \multicolumn{1}{c}{$R^2$} & \multicolumn{1}{c}{MAE} & \multicolumn{1}{c}{$R^2$} & \multicolumn{1}{c}{MAE} & \multicolumn{1}{c}{$R^2$} & \multicolumn{1}{c}{MAE} & \multicolumn{1}{c}{$R^2$} & ~ \\
         \midrule
         GWR & \underline{0.873} & 0.906 & 1.445 & 0.860 & 0.864 & 0.697 & 1.338 & 0.705 & 0.830 & 0.802 & \textbf{0.860} & \textbf{0.991} & \underline{0.807} & 0.607 & \underline{0.869} & 0.980 & 2 \\
         XGBoost & 0.957 & 0.895 & 1.719 & 0.820 & 0.891 & 0.746 & 1.642 & 0.649 & 0.840 & \underline{0.827} & 0.880 & \underline{0.991} & 0.822 & 0.720 & 0.912 & 0.979 & 0 \\
         SRGCNN & 2.805 & -0.271 & 2.904 & 0.374 & 1.78 & -0.184 & 1.754 & 0.611 & 1.013 & 0.691 & 0.982 & 0.984 & 1.066 & 0.499 & 0.900 & 0.977 &  0 \\
         GCNNWR & 0.897 & 0.907 & \textbf{1.133} & \textbf{0.929} & 0.898 & 0.768 & 1.013 & 0.858 & \underline{0.818} & 0.827 & \underline{0.868} & 0.990 & 0.864 & 0.713 & \textbf{0.857} & \textbf{0.983} & 4 \\
         GSH-EL & 1.382 & 0.720 & 2.884 & 0.157 & 0.961 & 0.647 & 2.178 & -0.178 & 1.115 & 0.547 & 2.938 & 0.828 & 0.917 & 0.558 & 1.791 & 0.921 & 0 \\
         \hline
         \makecell{Vanilla-mini} & 0.887 & 0.909 & \underline{1.213} & \underline{0.919} & \textbf{0.839} & 0.753 & 0.929 & \underline{0.884} & 0.913 & 0.800 & 1.124 & 0.984 & 0.836 & 0.709 & 0.995 & 0.978 & 1 \\
         \makecell{Vanilla-small} & 0.892 & 0.908 & 2.351 & 0.694 & \underline{0.844} & 0.751 & \textbf{0.897} & \textbf{0.892} & 0.878 & 0.817 & 0.980 & 0.988 & 0.833 & 0.710 & 0.924 & \underline{0.981} & 2 \\
         \makecell{Vanilla-large} & 0.880 & \underline{0.912} & 2.555 & 0.642 & 0.848 & 0.749 & 1.554 & 0.666 & 0.883 & 0.816 & 1.223 & 0.982 & 0.830 & 0.713 & 1.143 & 0.970 & 0 \\
         GA-mini & \textbf{0.870} & \textbf{0.920} & 1.269 & 0.911 & 0.881 & 0.751 & 0.946 & 0.876 & \textbf{0.818} & 0.810 & 1.039 & 0.985 & \textbf{0.804} & 0.710 & 1.054 & 0.971 & 4 \\
         GA-small & 0.905 & 0.910 & 1.530 & 0.840 & 0.872 & \textbf{0.771} & \underline{0.920} & 0.880 & 0.851 & \textbf{0.827} & 1.000 & 0.988 & 0.819 & \underline{0.736} & 1.280 & 0.964 & 2 \\
         GA-large & 0.905 & 0.911 & 2.383 & 0.646 & 0.870 & \underline{0.771} & 1.847 & 0.547 & 0.867 & 0.823 & 1.169 & 0.984 & 0.820 & \textbf{0.737} & 1.165 & 0.970 & 1 \\
         \bottomrule
         
    \end{tabular}
    \caption{Performance on our eight synthetic datasets. GeoAggregators (GA) achieved seven best results in total.}
    \label{tab:my_label}
\end{table*}

\subsection{Performance on Real-World Datasets}
We further compare the models on 3 representative real-world datasets, the results are summarized in Table 3.
On the smallest dataset (PM25), our GeoAggregator models show less satisfactory results than the best-performing model. This is because long-range dependence contributes less to the target point, considering the well-known complex spatial heterogeneity pattern in mainland China  \cite{wang2016measure}. We also observe that different data splits lead to significant variations in the results.

On the Housing price dataset, two GA-mini models achieve SOTA, indicating their efficiency in modeling densely distributed data points with governing SA and the ability to capture SH and global spatial structure through local modeling. XGBoost demonstrates its flexibility as a strong baseline in handling tabular data.

On the Poverty dataset, GA-small achieves the lowest MAE and a second-best {$R^2$} because it learns complex neighboring interactions governed by SA through inducing points. SRGCNN, GCNNWR and GSH-EL fail to complete training due to high memory requirements or prohibitively long runtime. XGBoost still performs well due to its ability to learn complex patterns and high flexibility.

\begin{table}
    \centering
    \small
    \renewcommand{\arraystretch}{0.95}
    \setlength{\tabcolsep}{3pt}
    \begin{tabular}{cccccccc}
    \toprule
        \multirow{2}*{Model} & \multicolumn{2}{c}{PM25} & \multicolumn{2}{c}{Housing} & \multicolumn{2}{c}{Poverty} & \multirow{2}*{\# best} \\
        \cmidrule(lr){2-3}
        \cmidrule(lr){4-5}
        \cmidrule(lr){6-7}
        ~ & \multicolumn{1}{c}{MAE} & \multicolumn{1}{c}{$R^2$} & \multicolumn{1}{c}{MAE} & \multicolumn{1}{c}{$R^2$} & \multicolumn{1}{c}{MAE} & \multicolumn{1}{c}{$R^2$} & ~ \\
        \hline
        GWR & 3.933 & 0.801 & 0.733 & 0.803 & 4.214 & 0.739 & 0 \\
        XGBoost & 4.017 & 0.813 & 0.645 & 0.888 & 3.622 & \textbf{0.845} & 1\\
        SRGCNN & 4.181 & 0.771 & 1.370 & 0.429 & - & - & 0 \\
        GCNNWR & \textbf{3.797} & \underline{0.832} & 0.704 & 0.895 & - & - & 1\\
        GSH-EL & \underline{3.863} & \textbf{0.842} & 0.718 & 0.860 & - & - & 1 \\
        \hline
        \makecell{Vanilla-mini} & 4.075 & 0.832 & \textbf{0.623} & \underline{0.906} & 3.563 & 0.839 & 1 \\
        \makecell{Vanilla-small} & 4.649 & 0.776 & 0.647 & 0.904 & 3.698 & 0.832 & 0 \\
        \makecell{Vanilla-large} & 4.494 & 0.787 & 0.649 & 0.900 & 3.637 & 0.835 & 0 \\
        GA-mini & 4.480 & 0.821 & \underline{0.624} & \textbf{0.911} & \underline{3.547} & 0.842 & 1 \\
        GA-small & 4.928 & 0.772 & 0.650 & 0.904 & \textbf{3.537} & \underline{0.844} & 1 \\
        GA-large & 4.721 & 0.778 & 0.641 & 0.896 & 3.565 & 0.842 & 0\\
        \bottomrule
    \end{tabular}
    \caption{Performance on three real-world datasets. GeoAggregators (GA) achieve 2 best results in total.}
    \label{tab:my_label}
\end{table}

\subsection{Computational Efficiency}
We compare the computational efficiency through the number of learnable parameters (\#Param) and number of floating operations (\#FLOPs) in one inference of each model (Table 4). As the dataset size increases, \#Param and \#FLOPs for GeoAggregators remain relatively stable, whereas other deep learning models exhibit a significant increase in both \#Param and \#FLOPs. The GeoAggregator models using the MCPA mechanism also demonstrate notably lower \#Param and \#FLOPs compared to those with vanilla attention. Finally, with an increasing number of attention layers, the \#Param and \#FLOPs values of GeoAggregators are effectively managed due to the introduction of inducing points in processor modules. The above characteristics of the GeoAggregator model make it well suited for efficient modeling across datasets of varying sizes.
\begin{table*}
    \centering
    \small
    \renewcommand{\arraystretch}{0.95}
    \setlength{\tabcolsep}{3pt}
    \begin{tabular}{crrrrrrrrr}
    \toprule
        \multirow{2}*{Model} & \multicolumn{2}{c}{Synthetic} & \multicolumn{2}{c}{PM25} & \multicolumn{2}{c}{Housing} & \multicolumn{2}{c}{Poverty} \\
        \cmidrule(lr){2-3}
        \cmidrule(lr){4-5}
        \cmidrule(lr){6-7}
        \cmidrule(lr){8-9}
         ~ & \multicolumn{1}{c}{\#Params} & \multicolumn{1}{c}{\#FLOPs} & \multicolumn{1}{c}{\#Params} & \multicolumn{1}{c}{\#FLOPs} & \multicolumn{1}{c}{\#Params} & \multicolumn{1}{c}{\#FLOPs} & \multicolumn{1}{c}{\#Params} & \multicolumn{1}{c}{\#FLOPs} \\
         \hline
        GWR & - & - & - & - & - & - & - & - & \\
        XGBoost & - & - & - & - & - & - & - & - & \\
        SRGCNN &  &  & 224.4K & 670.9M & 2820K & 8.5M & - & - \\
        GCNNWR & 1930K & 474.6M & 1210K & 231.2M & 8910K & 3281.8M & - & - \\
        GSH-EL & 460K & 29.3M & 220K & 14.0M & 2980K & 190.1M & - & - \\
        \hline
        \makecell{Vanilla-mini} & 7.6K & 1.7M & 7.9K & 3.5M & 7.9K & 3.6M & 8.2K & 3.7M \\
        \makecell{Vanilla-small} & 12.9K & 3.3M & 13.2K & 6.7M & 13.2K & 6.7M & 13.5K & 7.0M \\
        \makecell{Vanilla-large} & 18.2K & 10.5M & 18.4K & 21.3M & 18.5K & 21.4M & 18.8K & 21.8M \\
        GA-mini & \textbf{4.3K} & \textbf{0.7M} & \textbf{4.6K} & \textbf{1.6M} & \textbf{4.6K} & \textbf{1.6M} & \textbf{4.9K} & \textbf{1.7M} \\
        GA-small & \underline{6.3K} & \underline{1.4M} & \underline{6.5K} & \underline{2.9M} & \underline{6.6K} & \underline{3.0M} & \underline{6.8K} & \underline{3.2M} \\
        GA-large & 8.2K & 2.7M & 8.5K & 5.7M & 8.5K & 5.8M & 8.8K & 6.2M \\
        \bottomrule
    \end{tabular}
    \caption{The computational complexity of models in comparison, measured by \#Params and \#FLOPs.}
    \label{tab:my_label}
\end{table*}

\subsection{Ablation Study}
\subsubsection{Effect of the Attention Bias Factor.}
We conduct a series of experiments using three variants of the GeoAggregator on the Housing dataset with varying attention bias factor {$\lambda$}, as shown in Figure 3. Specifically, we fix {$\lambda$} as {$\{10^{-3},0.1,0.5,1,5,10,50,10^3\}$} during the training and testing stage. All 3 GeoAggregators perform not well when {$\lambda=10^{-3}$} or {$\lambda=10^{3}$}, i.e., when little or too much geographical prior is added to guide the local attention. In contrast, 3 GeoAggregators perform best when using a balanced {$\lambda$}, demonstrating the validity of introducing the Gaussian bias. We also marked the results of 3 variants with learnable {$\lambda$} with red stars in Figure 3. The learned {$\lambda$}s are consistent with our ablation experiments, further indicating the effectiveness of our design.

\begin{figure}
    \centering
    \includegraphics[width=1\linewidth]{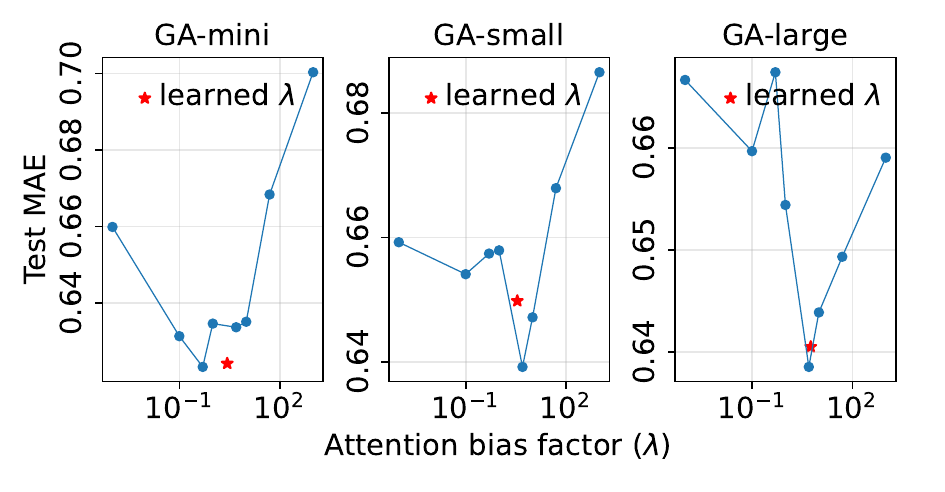}
    \caption{Effect of the attention bias factor {$\lambda$} of 3 variants of GeoAggregator, on the Housing dataset.}
    \label{fig:enter-label}
\end{figure}

\subsubsection{Effect of the Sequence Length.}
GeoAggregator models selectively capture neighboring interactions through a local attention operation. We train a GeoAggregator-mini model (without the processor module) with varying input sequence lengths {$\ell_{max}$} to assess the impact of sequence length on the performance. As shown in Figure 4, the performance (in terms of test MAE) continues to improve with increasing sequence length until it reaches 0.618 when {$\ell_{max}=1024$}. This indicates that the GeoAggregator learns to efficiently aggregate useful information from local input sequences. Unlike GWR or graph-based models, GeoAggregator eases the over-smoothing problem when the bandwidth or neighborhood size is overly large. That is, it's robust to less relevant data points, even within a large neighborhood.

Based on the previous discussion, we argue that for datasets with unclear SA and SH patterns, one can gradually increase {$\ell_{max}$} until one reaches a satisfactory result that balances computational efficiency and prediction accuracy.

\begin{figure}
    \centering
    \includegraphics[width=1\linewidth]{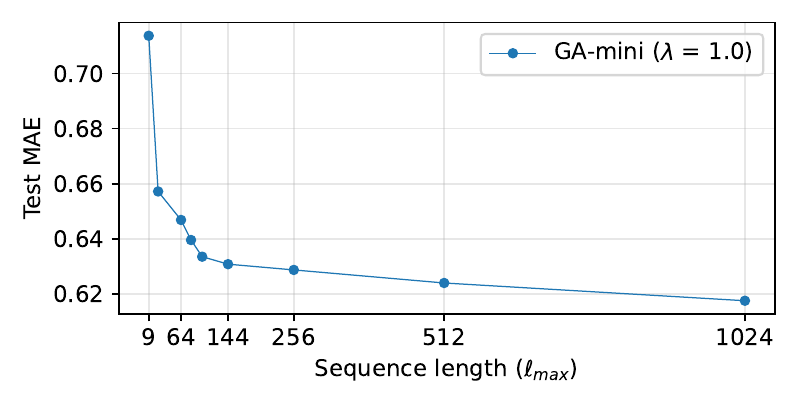}
    \caption{Effect of the input sequence length {$\ell_{max}$}. We compare results of the GA-mini model with {$\lambda = 1.0$}, on the Housing dataset.}
    \label{fig:enter-label}
\end{figure}

\subsubsection{Computational Complexity of Different Attention Mechanisms.}
To demonstrate the computational efficiency of the proposed Multi-head Cartesian Product Attention (MCPA), we compare the number of floating-point operations (\#FLOPs) in one inference of the vanilla full attention, vanilla inducing point attention, and our MCPA on the Housing dataset. All 3 models contain 3 attention layers, with the inducing point models incorporating 4 inducing points in both the first and second attention layers.

We report \#FLOPs with increasing {$\ell_{max}$} in Figure 5. It is depicted that by introducing inducing points, the computational complexity of the transformer architecture scales linearly with growing {$\ell_{max}$}. Moreover, replacing the vanilla attention mechanism with our MCPA further reduces the computational complexity, making it suitable for handling large datasets that involve dense point distributions and complex geospatial effects.

\begin{figure}
    \centering
    \includegraphics[width=1\linewidth]{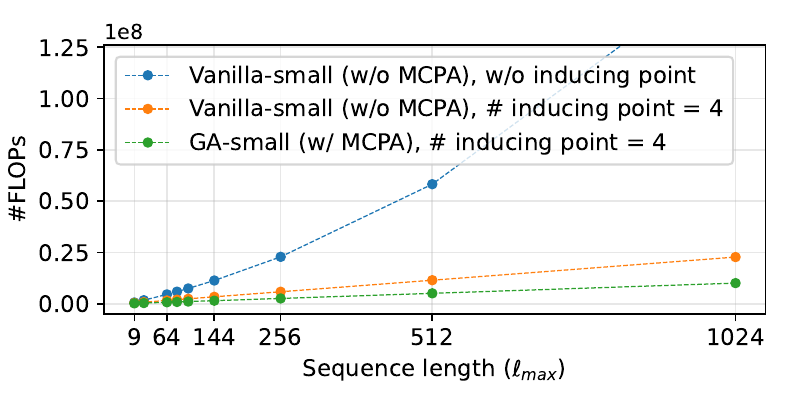}
    \caption{Computational cost of one inference of different attention mechanisms, on the Housing dataset.}
    \label{fig:enter-label}
\end{figure}

\section{Conclusion and Future Work}
In this work, we address the issues of scalability and flexibility in current deep networks for geospatial tabular data modeling, by introducing a novel lightweight transformer-based model, GeoAggregator. GeoAggregator explicitly accounts for spatial autocorrelation and spatial heterogeneity effects through a novel Gaussian-biased Cartesian product attention mechanism and a global positional awareness. Our GeoAggregator model shows superior performance and computational efficiency on synthetic and real-world datasets compared to several baseline models, offering a promising solution for geospatial tabular data tasks. 

For future work, tailoring the bias for each spatial feature individually could potentially improve performance. In addition, incorporating new input heads to handle categorical variables would also be a beneficial improvement.

\bibliography{aaai25}

\appendix
\renewcommand{\thefigure}{A\arabic{figure}}
\setcounter{figure}{0}

\newpage

\section{Appendix}
\subsection{Generation of Synthetic Datasets}
\subsubsection{Data Generation Process}
For the synthetic datasets, we consider 4 types of data generation processes (DGPs). For the Spatial Linear Model (Lin), for each point {$\boldsymbol{p}_i$}, the target variable is a simple linear combination of 2 covariates {$\mathbf{x}_1$} and {$\mathbf{x}_2$}:

\begin{align}
    \mathbf{y} = \beta_0 + \beta_1 \mathbf{x}_1 + \beta_2 \mathbf{x}_2 + \mathbf{\epsilon}
\end{align}

where {$\mathbf{\epsilon}$} is an error term. In our experiments, we sample the error from a Gaussian distribution: {$\epsilon_i \sim N(0,1)$}, independent to spatial locations.

In the Spatial Lagged Model (SL), spatially lagged target variables are considered on top of the Lin process:

\begin{align}
    \mathbf{y} = \rho \mathbf{W} \mathbf{y} + \beta_0 + \beta_1 \mathbf{x}_1 + \beta_2 \mathbf{x}_2 + \mathbf{\epsilon}
\end{align}

where {$\mathbf{y} \in \mathbb{R}^N$} is a vector containing {$N$} data points, {$\rho$} is a parameter controlling the strength of spatial autocorrelation (SA). {$W$} is a spatial weight matrix, specifying the pattern and degree of interactions between each point and its other points. In this paper, we propose to use a binary matrix {$\mathbf{W} \in \{0,1\}^{N \times N}$} ({$1$} for valid contiguity of corresponding pair of points) defined by the Queen's adjacency rule.

Another process, Spatial Lagged X Model (SLX), introduces spatially lagged covariates:

\begin{align}
    \mathbf{y} = \theta_1 \mathbf{W}_1 \mathbf{x}_1 + \theta_2 \mathbf{W}_2 \mathbf{x}_2 + \beta_0 + \beta_1 \mathbf{x}_1 + \beta_2 \mathbf{x}_2 + \mathbf{\epsilon}
\end{align}

Incorporating both spatially lagged covariates and the target variable, we have the Spatial Durbin Model (Durbin):

\begin{align}
    \mathbf{y} = \theta_1 \mathbf{W}_1 \mathbf{x}_1 + \theta_2 \mathbf{W}_2 \mathbf{x}_2 + \rho \mathbf{W}_3 \mathbf{y} + \beta_0 + \beta_1 \mathbf{x}_1 + \beta_2 \mathbf{x}_2 + \mathbf{\epsilon}
\end{align}

In this paper, we use the same regression coefficients {$\beta_{0,1,2}$} for all 4 DGPs, as proposed in \cite{li2022extracting} (see Figure A1). Different DPGs are used to exam whether the spatial regression models could capture varying SA and SH effects.

\begin{figure}
    \centering
    \includegraphics[width=1\linewidth]{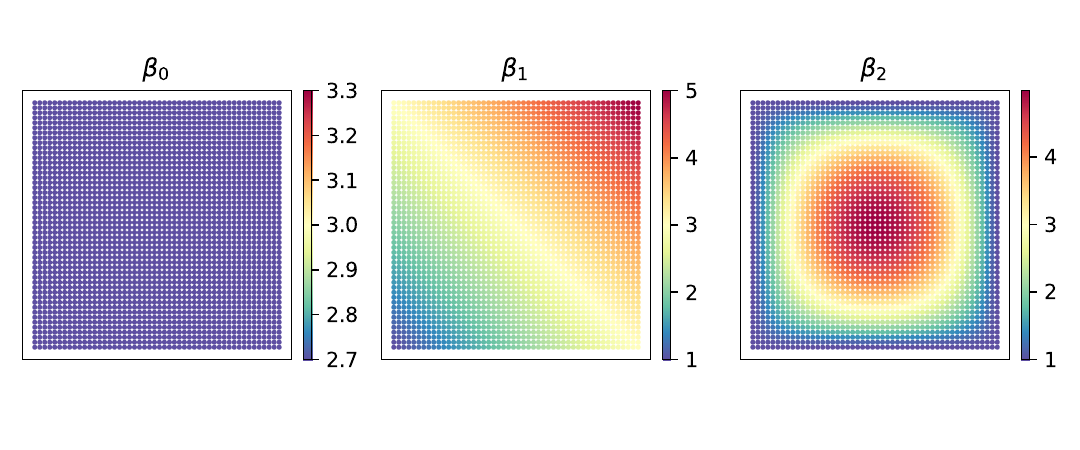}
    \caption{Maps of regression coefficients ({$\beta$}s) used in the DGPs.}
    \label{fig:enter-label}
\end{figure}

\subsubsection{Map Visualization of Synthetic Datasets}

As mentioned in the Experiments section, we use two types of covariates as input of the DGPs. They are sampled from a spatially independent uniform distribution and a real-world Digital Elevation Map (DEM), respectively (as seen in Figure A2). Target variables are generated from DGPs accordingly (Figure A3).

\begin{figure}
    \centering
    \includegraphics[width=1\linewidth]{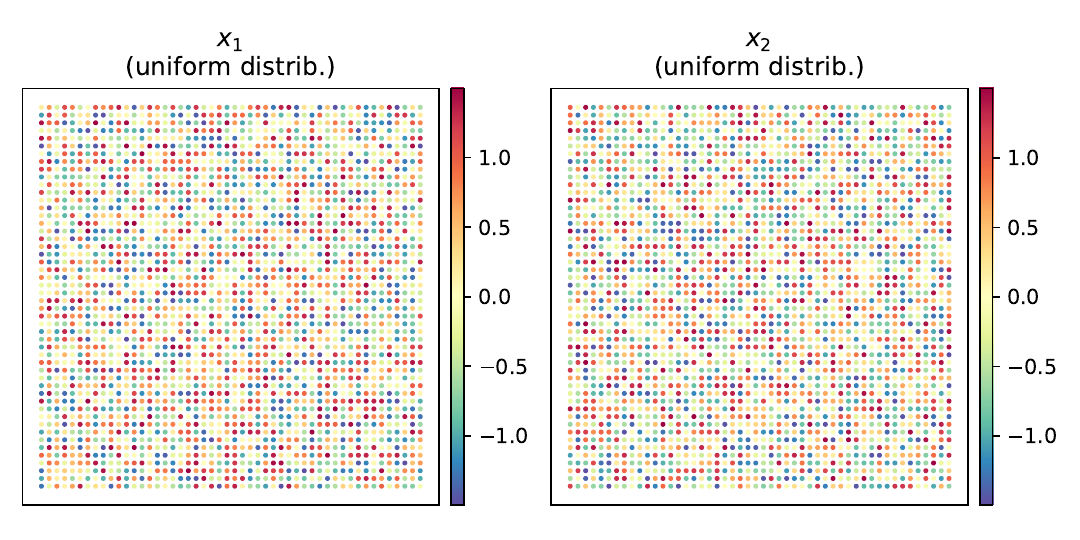}
    \includegraphics[width=1\linewidth]{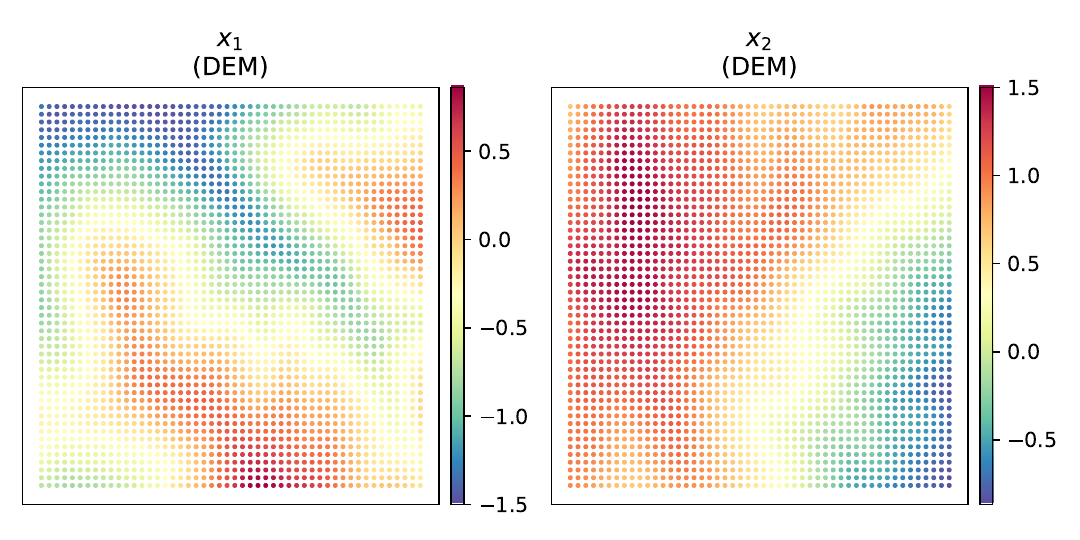}
    \caption{Maps of two types of covariates used in this paper. The first line shows covariates randomly sampled from a uniform distribution, showing spatial stationary. The second line illustrates covariates sampled from a real-world Digital Elevation Map (DEM) data, showing spatial non-stationary.}
    \label{fig:enter-label}
\end{figure}

\begin{figure}
    \centering
    \includegraphics[width=1\linewidth]{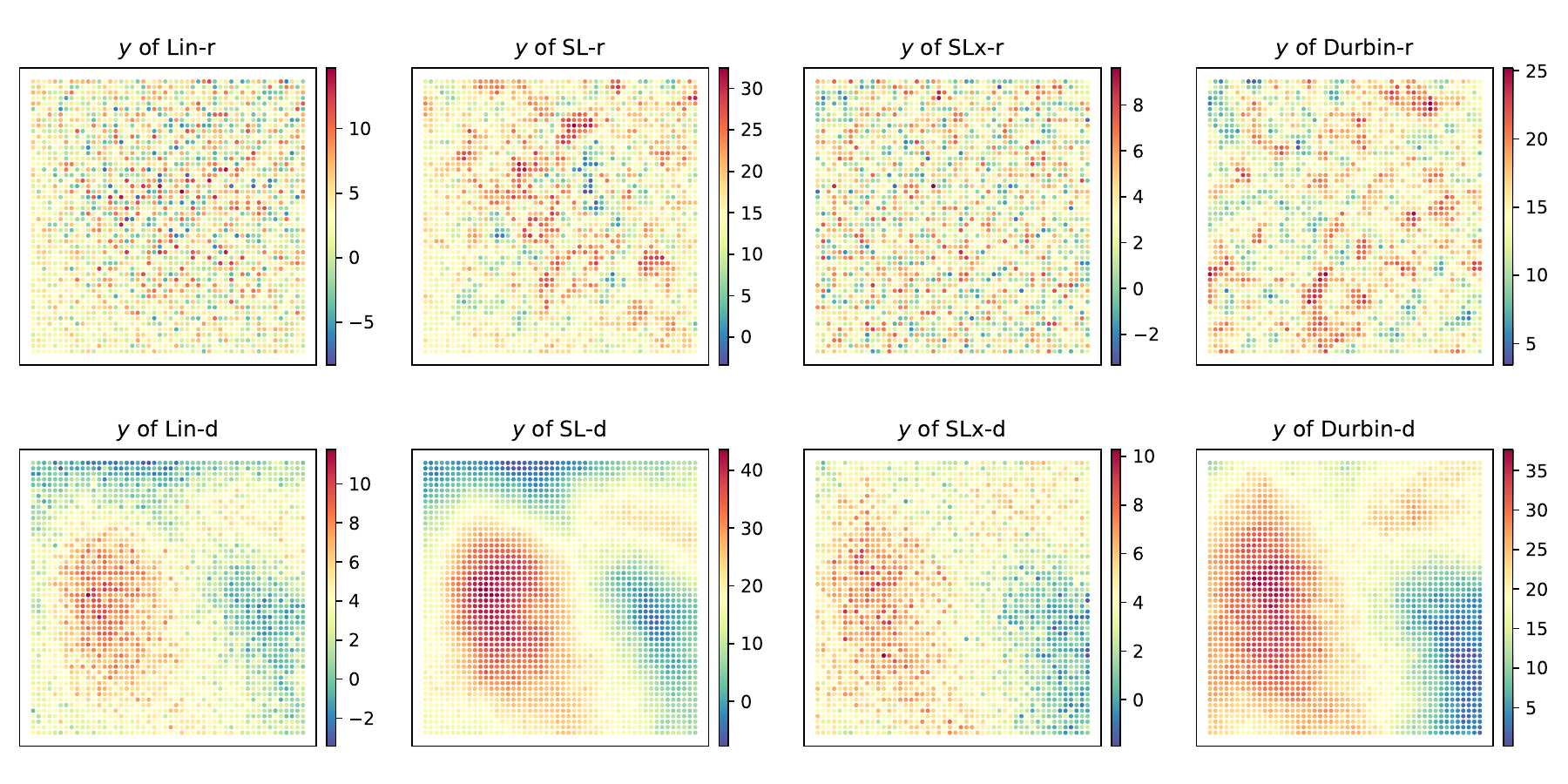}
    \caption{Maps of the generated target variables.}
    \label{fig:enter-label}
\end{figure}

\subsection{Map Visualization of Real-World Datasets}

3 real-world datasets (PM25, Housing and Poverty) used in this paper is visualized in Figure A4, A5 and Figure A6, respectively. Different distribution patterns and the range of the spatial effect are illustrated.

\begin{figure}
    \centering
    \includegraphics[width=1\linewidth]{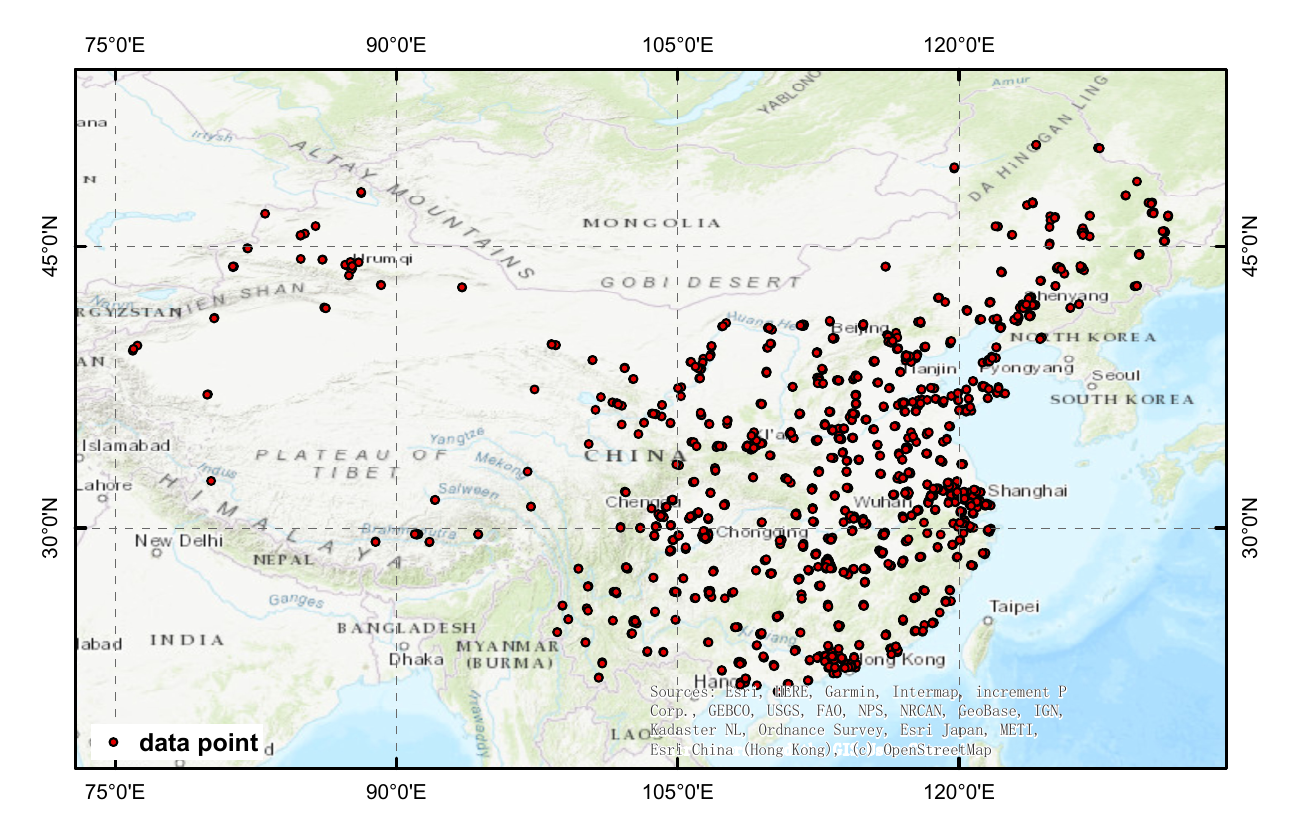}
    \caption{Mapping of the PM25 dataset \cite{DaiPm25CNN}. Note that the distribution of data points is relatively sparse and uneven across the mainland of China.}
    \label{fig:enter-label}
\end{figure}

\begin{figure}
    \centering
    \includegraphics[width=1\linewidth]{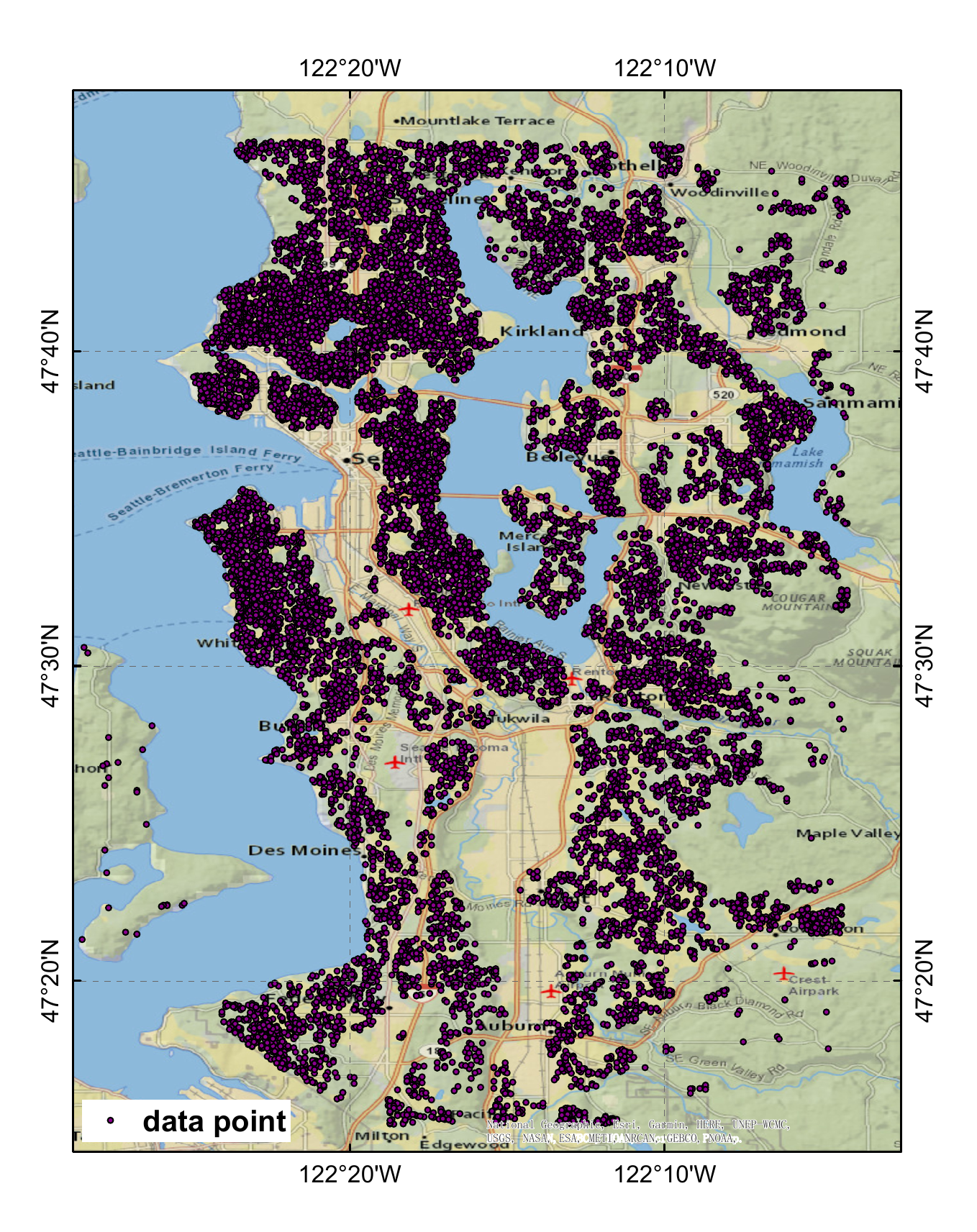}
    \caption{Mapping of the Housing dataset \cite{li2024geoshapley}. Note that points are densely distributed on a city block scale.}
    \label{fig:enter-label}
\end{figure}

\begin{figure}
    \centering
    \includegraphics[width=1\linewidth]{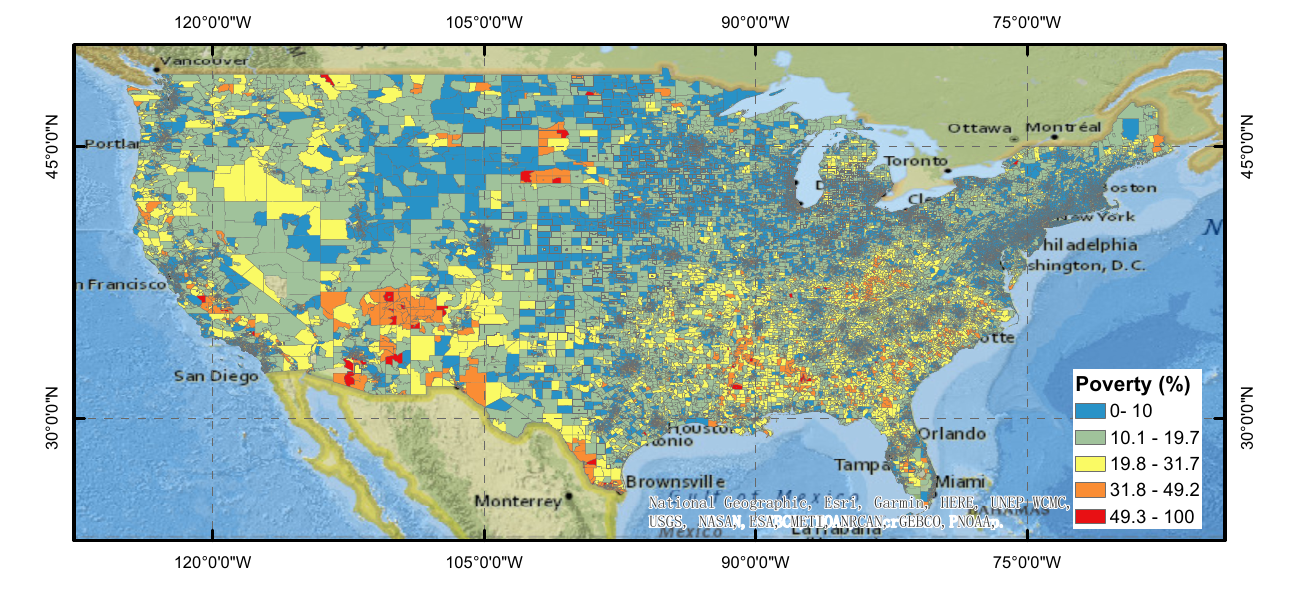}
    \caption{Mapping of the Poverty dataset \cite{MaryniaSDOH}. Note that data points are the centroid of the neighborhood polygons. Data points are densely distributed across the continental US.}
    \label{fig:enter-label}
\end{figure}

\end{document}